\documentclass[letterpaper, 10 pt, conference]{ieeeconf}  

\IEEEoverridecommandlockouts                              

\renewcommand{\baselinestretch}{0.99}

\usepackage{graphicx}
\usepackage{booktabs}
\usepackage{tabularx}
\usepackage{todonotes}
\usepackage{algorithm}
\usepackage{algorithmicx,algpseudocode}

\usepackage{multicol}
\usepackage[bookmarks=true]{hyperref}
\usepackage{multirow}
\usepackage{times}
\usepackage{epsfig}
\usepackage{graphicx}
\usepackage{amsmath}
\usepackage{amssymb}
\makeatletter
\let\NAT@parse\undefined
\makeatother
\usepackage[numbers]{natbib}

\usepackage{booktabs}
\usepackage{tabularx}
\usepackage[font=small,labelfont=bf]{caption}

\usepackage[font=small]{caption}

\usepackage{hyperref}
\usepackage[capitalise, nameinlink]{cleveref}

\title{\LARGE \bf Bridging Speech, Emotion, and Motion: a VLM-based Multimodal Edge-deployable Framework for Humanoid Robots}

\author{
    Songhua Yang$^1$, Xuetao Li$^1$, Xuanye Fei$^1$, Mengde Li$^1$, Miao Li$^{1*}$%
    \thanks{1. School of Computer Science, Wuhan University}
    \thanks{* Corresponding Author}
}

\usepackage{fancyhdr}
\begin{document}
\maketitle

\begin{abstract}
Effective human-robot interaction requires emotionally rich multimodal expressions, yet most humanoid robots lack coordinated speech, facial expressions, and gestures. Meanwhile, real-world deployment demands on-device solutions that can operate autonomously without continuous cloud connectivity.
To bridging \underline{\textit{S}}peech, \underline{\textit{E}}motion, and \underline{\textit{M}}otion, we present \textit{SeM$^2$}, a Vision Language Model-based framework that orchestrates emotionally coherent multimodal interactions through three key components: a multimodal perception module capturing user contextual cues, a Chain-of-Thought reasoning for response planning, and a novel Semantic-Sequence Aligning Mechanism (SSAM) that ensures precise temporal coordination between verbal content and physical expressions. 
We implement both cloud-based and \underline{\textit{e}}dge-deployed versions (\textit{SeM$^2_e$}), with the latter knowledge distilled to operate efficiently on edge hardware while maintaining 95\% of the relative performance. Comprehensive evaluations demonstrate that our approach significantly outperforms unimodal baselines in naturalness, emotional clarity, and modal coherence, advancing socially expressive humanoid robotics for diverse real-world environments.
\end{abstract}

\section{INTRODUCTION}

Humanoid robots recently attracted considerable interest from both industry and academia due to their intrinsic embodiment and anthropomorphic configuration in the domain of human-robot interaction (HRI) \cite{Ottoni2021ARO,brohan2023can,garcia2007evolution}. While substantial research efforts focused on developing robots' locomotive and manipulation capabilities, the equally crucial aspect of emotional expressivity received comparatively less attention. However, emotionally rich dialogue, facial expressions, and gesture cues constitute fundamental characteristics of human-like interaction that extend beyond mere functional operations. This emotional dimension is particularly critical as robots increasingly transition from controlled industrial environments to social spaces where they must engage with humans in nuanced, context-sensitive ways \cite{aly2013model,saunderson2019robots}.


Traditional approaches to HRI were predominantly relied on disparate models for different modalities, resulting in significant coordination challenges across speech, emotion, and motion \cite{takayama2011expressing,huang2012robot,kato2015may}. Rule-based and template-based systems, while functional for controlled environments, lack scalability when faced with the complexity of multimodal interactions \cite{porfirio2018authoring,leonardi2019trigger,li2020animated}. Even data-driven approaches using conventional machine learning techniques require specialized datasets for each behavioral type, making adaptation to novel scenarios prohibitively expensive \cite{huang2014learning,marmpena2019generating,sripathy2022teaching,oralbayeva2024data}.

The emergence of Large Language Models (LLMs), particularly Vision-Language Models (VLMs), offers a promising approach to effectively align diverse modalities in HRI \cite{wake2024gpt,wang2024vlm}. These models demonstrate remarkable capabilities in contextual understanding and content generation across different expressive modals. Initial applications of large language models to robotics have shown encouraging results in instruction parsing and task planning \cite{brohan2023can,huang2023voxposer,li2024manipllm}, yet their potential for multimodal emotional expression coordination remains largely unexplored.

To address this gap, we present \textbf{SeM$^2$} and \textbf{SeM$^2_e$}, a VLM-based framework designed to generate coherent, emotionally expressive multimodal interactions for humanoid robots. We introduce a novel Semantic-Sequence Aligning Mechanism (SSAM) module, which establishes semantic bridges between textual content and expressive actions through temporal constraint optimization, enabling precise synchronization of speech, facial expressions, and physical gestures. In addition, our framework offers two deployment options: a high-performance API-based solution that leverages cloud computing, and a knowledge-distilled edge deployment model that maintains core functionality while operating efficiently on resource-constrained hardware.

Our extensive experimental evaluation, involving both AI assessment and human expert ratings, demonstrates that the proposed framework significantly outperforms unimodal approaches across all metrics, particularly in emotional clarity and overall user experience. The edge-deployed model retains approximately 95\% of the performance of the API-based model while enabling real-time operation on embedded systems. Ablation studies further confirm the critical contribution of our SSAM component, with modal synchronization emerging as the most influential factor in interaction quality.

The contributions of our work are as follows: \textbf{(1)} we propose a novel VLM-based framework to generate· coherent multimodal expressions in humanoid robots; \textbf{(2)} we introduce the SSAM to establish precise temporal coordination between speech content and expressive behaviors; \textbf{(3)} we demonstrate the feasibility of deploying sophisticated multimodal interaction systems on edge devices through knowledge distillation, opening new possibilities for autonomous socially expressive robots in real-world applications.

\begin{figure*}[ht!]
    \centering
    \includegraphics[width=\textwidth]{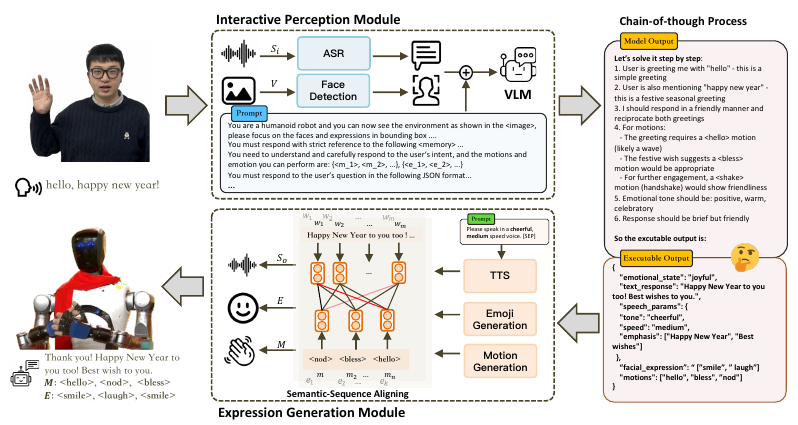}
    \caption{The framework of our SeM$^2$, the left side is a input and output of a specific example. 
    }
    \label{fig:process}
\end{figure*}

\section{RELATED WORK}

\subsection{Large Language Models for Robotics}



In recent years, LLMs sparked a new wave of AI technological innovation, achieving significant progress in robotic applications \cite{zhao2023survey,achiam2023gpt,liu2024deepseek,yang2024zhongjing}. Leveraging LLMs and VLMs' powerful comprehension, perception, and reasoning capabilities, robots can now more precisely parse instructions, perceive environments, and plan tasks \cite{zeng2023large,zhong2023can}. The emergence of VLM particularly provides robots with a unified multimodal and multitasking framework, enabling improved contextual understanding when processing multimodal information \cite{wake2024gpt,wang2024vlm}.

Currently, LLM applications in robotics focus primarily on manipulation tasks requiring instruction parsing and task planning. For instance, SayCan \cite{brohan2023can} employed LLM as a high-level task planner, combining perception to generate long-horizon operational strategies for robots. VoxPoser \cite{huang2023voxposer} synthesized adaptive robot trajectories using 3D value maps derived from large language models. ManipLLM \cite{li2024manipllm} utilized refined chain-of-thought reasoning to predict end-effector poses, integrating closed-loop impedance strategies for more stable generalized manipulation. RobotGPT \cite{jin2024robotgpt} constructed a stable and secure decision-making framework on ChatGPT's foundation, employing trained agents to replace direct control code generation.

Moreover, with the gradual emergence of the ``Foundation Models" concept, recent works have begun exploring the training of generalizable strategies using large-scale, multi-task, and multimodal datasets, aiming to endow robots with cross-task generalization capabilities \cite{brohan2023rt,liu2024rdt,kim2024openvla,zhao2025vlas}. These approaches represent a significant shift towards more adaptable and versatile robotic systems that can leverage comprehensive pre-training to improve performance across diverse scenarios.




\subsection{Expressive Human-Robot Interactions}

During the past decade, HRI research has expanded from focusing solely on functional tasks (e.g. grasping and manipulating) to encompass affective and expressive interactions \cite{Ottoni2021ARO,garcia2007evolution,li2014learning,gao2023two}. Along this line, the ability of robots to generate more natural, flexible, and multimodal behaviors became a core research challenge in robotics, which undergoes several evolutionary stages \cite{aly2013model,saunderson2019robots,brohan2023can}.

Early studies relied predominantly on rule-based or template-based methods \\ \cite{takayama2011expressing,porfirio2018authoring,leonardi2019trigger,li2020animated}. Rule-based approaches required manually predefining interaction rules and operational logic, such as performing a nod or greeting when detecting a user's smile, or extracting patterns from videos of observed human interaction, which were then codified into a set of fixed rules \cite{huang2012robot,kato2015may}. However, these manually configured approaches often struggled with scalability in complex multimodal interactions. Template-based approaches learned a set of basic interaction patterns from a small number of examples and reused them in subsequent scenarios, which offers better extensibility than rule-based approaches, but still faces limitations in generating diverse behaviors and efficiently coordinating multiple modalities \cite{ferrarelli2018design,porfirio2020transforming,oralbayeva2024data}.
With the evolution of AI paradigms, some works began adopting data-driven strategies that employ machine learning models to discover patterns in large-scale HRI datasets, thereby learning more adaptive robotic behaviors \cite{huang2014learning,liu2016data,marmpena2019generating,sripathy2022teaching}. 
Although these approaches have improved the generative capacity and diversity of robot behaviors, they typically rely on substantial training data and require collecting dedicated datasets for each type of behavior, making adaptation or extension to new scenarios costly \cite{oralbayeva2024data}.

Meanwhile, recent work has demonstrated the potential of generative models, particularly VLMs in HRI. For example, \cite{rodriguez2019spontaneous,suguitan2020moveae} integrated multimodal expressions into generative models to reduce the dependency on manual orchestration or large specialized datasets. Through only a small number of examples, \cite{zhou2018cost} synthesized diverse behaviors and iteratively optimized them based on user feedback in real time. GenEM \cite{mahadevan2024generative} used VLM to generate control sequences in dialog contexts using predefined skill functions, enabling robots to express motion and emotion. In \cite{huang2024emotion}, the in-context learning of VLM was used to generate expressive gestures suited to social settings in humanoid robots.

However, these works focus on a relatively simplified robot embodiment, offering limited multimodal emotional interaction and often overlooking a deeper emotional engagement with users \cite{mahadevan2024generative}. For more sophisticated humanoid robots, additional exploration is needed to align the speech, emotion, motion, etc., thus achieving more natural and expressive HRI.

\section{Methods}


We propose \textbf{SeM$^2$} and \textbf{SeM$^2_e$}, a VLM-based framework to generate expressive multimodal interactions in humanoid robots. The framework aims to enable natural, fluent and emotionally rich human-robot interactions, supporting both cloud and edge deployments, as illustrated in Figure \ref{fig:process}.

\subsection{Task Definition}


SeM$^2$ takes human speech instructions $S_i$ and simultaneous visual observations $V$ as input. The system outputs a verbal response $S_o$ and two modal sequences $E=\{e_1,e_2,...,e_m\}$ and $M=\{m_1,m_2,...,m_n\}$, representing facial expressions and action execution sequences, respectively. Formally, the system aims to learn a mapping function $f: (S_i, V) \rightarrow (S_o, E, M)$ that maintains coherence across three critical dimensions:

\begin{figure}[]
    \centering
    \includegraphics[width=0.45\textwidth]{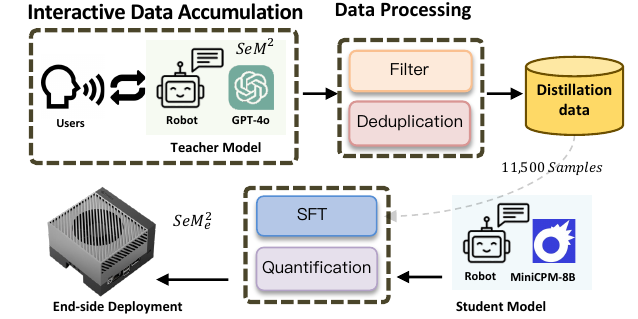}
    \caption{The detailed construction process of SeM$^2_e$ edge deployment. }
    \label{fig:end_side}
\end{figure}

\subsection{Multimodal Interactive Perception}


Rich perception is crucial for robots to identify and comprehend human communication cues \cite{sciutti2018guest}. Our SeM$^2$ interaction perception module extends beyond traditional visual and audio channels to capture more nuanced user language content, emotional states, and contextual information.

    
\begin{itemize}
    \item \textbf{Speech Perception and Processing:} The speech module must not only accurately understand the semantic content of user instructions but also capture the emotional nuances embedded in vocal intonation. We employ SenseVoice\footnote{\url{https://huggingface.co/FunAudioLLM/SenseVoiceSmall}} as our speech processing module, leveraging its exceptional multilingual support and emotion recognition capabilities critical for constructing emotionally rich human-robot interactions.
    
    \item \textbf{Facial Detection:} Precise facial localization enables the VLM to more accurately analyze user expressions and emotions, significantly enhancing the robot's understanding of user states. We utilize the YOLOv8-face model \cite{vemulapalli2023face} for user facial detection and localization.
\end{itemize}

These multimodal perception results, combined with carefully designed prompt templates, are input into the VLM, providing comprehensive interaction information for subsequent chain-of-thought reasoning and response generation.

\subsection{Chain-of-Thought Prompt}


Rich perception is crucial for robots to identify and comprehend human communication cues \cite{sciutti2018guest}. Our SeM$^2$ interaction perception module extends beyond traditional visual and audio channels to capture more nuanced user language content, emotional states, and contextual information.

    

\begin{itemize}
    \item \textbf{Multimodal Perception Integration:} Guide to the model to comprehensively analyze visual inputs $V$ and speech $S_{i}$. Structured prompts enable unified parsing of user voice content, facial expressions, and body postures, forming a holistic perception of the user's current state. The prompt also incorporates a specific knowledge base as prior information to ensure appropriate and persona-consistent responses.
    
    \item \textbf{Modal Expression Coordination:} Requiring explicit consideration of semantic consistency and emotional coordination between language, facial expressions and actions. Our CoT prompt instructs the model to consider the relationship between modalities when generating each expression, such as ensuring coordinated expression combinations (e.g., language content, facial changes, and gestures) when expressing surprise, effectively avoiding inter-modal expression conflicts.
    
    \item \textbf{Embodiment Constraint Awareness:} By explicitly specifying executable action and expression script sets $M=\{m_1, m_2, ...\}$, $E=\{e_1, e_2, ...\}$, the model can recognize and adhere to the robot's physical constraints, generating executable action and expression commands.
\end{itemize}

\subsection{Multimodal Expression Generation}


The multimodal expression generation module is responsible for transforming the VLM's structured output into executable robotic behaviors, implementing the mapping process $g: (S_o, E, M) \rightarrow A$, where $A=\{A_e, A_m\}$ represents the robot's actual set of facial expressions and actions. Leveraging the meticulously designed CoT Prompt, the VLM performs multidimensional deep reasoning and outputs expression plans in a structured JSON format. As shown in the right-hand side of Figure \ref{fig:process}, the system sends parsed instructions to three parallel execution modules: speech synthesis, facial expression generation, and motion control. We employ ChatTTS\footnote{\url{https://huggingface.co/2Noise/ChatTTS}} for speech synthesis, which allows direct control of emotional parameters and speed through prompts, while facial expressions and actions are mapped to predefined executable script sequences by the VLM.

\subsubsection{Semantic-Sequence Aligning Mechanism}



Although VLM-generated action and expression sequences are semantically related to textual content, they typically lack precise temporal alignment, often resulting in robotic expressions with inappropriate timing or rhythmic inconsistencies. In particular, for long vocal content or complex action sequences, simple parallel triggering fails to meet the naturalness requirements of human-robot interaction. To address this challenge, we propose the Semantic-Sequence Aligning Mechanism (SSAM), illustrated in the lower part of Figure \ref{fig:process}. The core concept of SSAM is to establish a semantic bridge between textual words and expressive actions by solving a temporal constraint optimization problem via dynamic programming, thereby determining the optimal execution plan.

Initially, we tokenize the speech text $S_o = \{w_1, w_2, ..., w_n\}$ and perform a time mapping to estimate the temporal position of each word during speech synthesis:

\begin{equation}
\begin{aligned}
t^s_i &= \sum_{k=1}^{i-1} \tau(w_k) \cdot \alpha(\texttt{speed}) \\
t^e_i &= t^s_i + \tau(w_i) \cdot \alpha(\texttt{speed})
\end{aligned}
\end{equation}

Here, $\tau(w_k)$ represents the baseline duration of word $w_k$, and $\alpha(\texttt{speed})$ is an adjustment factor based on speech rate parameters. Next, we calculate the semantic relevance between words and actions/expressions using pre-trained embeddings:

\begin{equation}
S(w_i, a_j) = \cos(\text{Emb}(w_i), \text{Emb}(a_j))
\end{equation}

Where $a_j$ represents expression or action identifiers, and $\text{Emb}(\cdot)$ is a pre-trained word embedding model. By setting a threshold $\theta$, the system filters word-action pairs with significant relevance. Finally, the system solves a temporal constraint optimization problem to determine the optimal execution timing for each expressive action:

Where $a_j$ represents expression or action identifiers, and $\text{Emb}(\cdot)$ is the \\ \texttt{BERT-base-uncased}\footnote{\url{https://huggingface.co/google-bert/bert-base-uncased}} model (hidden dimension 768). By setting a threshold $\theta=0.7$ (determined by a validation set to balance precision and recall), the system filters the word-action pairs with significant relevance. Finally, the system solves the temporal constraint optimization problem using dynamic programming to determine the optimal execution timing for each expressive action:

\begin{equation}
\begin{aligned}
\max_{T} & \sum_{j=1}^{m} \max_{i} \{S(w_i, a_j) \cdot \mathbb{I}(|T(a_j) - t^s_i| < \delta)\} \\
\text{subject to:} \\
& T(a_j) + d(a_j) \leq T(a_{j+1}), \forall j \in \{1,2,...,m-1\} \\
& |T(a_j) - T(a_k)| > \max(d(a_j), d(a_k)), \forall (a_j, a_k) \in \mathcal{C}
\end{aligned}
\end{equation}

Here, $T(a_j)$ represents the execution time of action $a_j$, $d(a_j)$ its estimated duration, $\delta$ the range of allowed time deviations, and $\mathcal{C}$ the set of conflicting action pairs.

Here, $T(a_j)$ represents the execution time of action $a_j$, $d(a_j)$ its estimated duration (e.g., $d(<hello>'')=1.2$s, $d(<nod>)=0.8$s), $\delta=0.3$s the range of allowed time deviations (empirically set to match human perception tolerance), and $\mathcal{C}$ the set of conflicting action pairs (e.g., $\mathcal{C}$ includes ``$<wave>$'', ``$<handshake>$'' as they involve overlapping arm movements). The dynamic programming solution iteratively selects the optimal $T(a_j)$ by maximizing semantic similarity while satisfying temporal constraints, ensuring globally consistent alignment.

SSAM leverages semantic similarity to establish a natural connection between content and expression, precisely synchronizing key actions with related vocal content and enhancing interaction naturalness and expressiveness. Temporal constraints ensure the physical feasibility of expression sequences, avoiding action conflicts. For example, in the phrase ``Happy New Year to you too!'', SSAM successfully aligns the ``$<hello>$'' action with ``Happy'' (synchronized within $\delta$), synchronizes the ``$<bless>$'' action with ``New Year'', and coordinates the ``$<nod>$'' action with the closing, all validated via motion capture data to match human interaction rhythms.



\subsubsection{The Edge-deployable Model using Knowledge Distillation}
As mobile interaction terminals, humanoid robots must maintain stable interaction capabilities in unpredictable environments where cloud VLM APIs can compromise user experience through network latency and connection interruptions. To address this, we constructed an edge-deployable model \textbf{SeM$^2_e$} using knowledge distillation \cite{xu2024survey}, as illustrated in Figure \ref{fig:end_side}. We implemented SeM$^2$ with GPT-4o as the teacher model, accumulating 52,000 raw interaction samples covering diverse scenarios, and used text semantic hashing (SimHash) to filter them into 11,500 high-quality examples with a duplication rate of $>$ 70\%.

For edge deployment, we selected MiniCPM-8B as the student model backbone and employed supervised fine-tuning to transfer the teacher model's capabilities. The model was then INT4-quantized ($\hat{W} = \text{clip}(\lfloor\frac{W}{\Delta}\rceil, -8, 7) \cdot \Delta$) for deployment on NVIDIA Jetson Orin, significantly reducing computational requirements while maintaining core functionalities and achieving real-time interaction performance on resource-constrained hardware.

\begin{table*}[h!]
\centering
\caption{The AI expert evaluation main results (5-point likert scale, N=5), $±$ represents the standard deviation.
}
\label{tab:main_results}
\resizebox{0.8\textwidth}{!}{%
\renewcommand{\arraystretch}{1.2}
\begin{tabular}{@{}llccccc@{}}
\toprule
\textbf{System} & \textbf{Version} & \textbf{Naturalness} & \textbf{Emotional} & \textbf{Modal} & \textbf{Response} & \textbf{Overall} \\
& & & \textbf{Clarity} & \textbf{Coherence} & \textbf{Appropriateness} & \textbf{Experience} \\
\midrule
\multirow{2}{*}{Full} & SeM$^2$ & 4.42 $\pm$ 0.21 & 4.51 $\pm$ 0.18 & 4.38 $\pm$ 0.22 & 4.45 $\pm$ 0.19 & 4.53 $\pm$ 0.17 \\
& SeM$^2_e$ & 4.28 $\pm$ 0.24 & 4.33 $\pm$ 0.22 & 4.21 $\pm$ 0.25 & 4.29 $\pm$ 0.23 & 4.37 $\pm$ 0.20 \\
\midrule
\multirow{2}{*}{No-Expr} & SeM$^2$ & 3.72 $\pm$ 0.25 & 3.24 $\pm$ 0.27 & 3.96 $\pm$ 0.24 & 4.39 $\pm$ 0.21 & 3.78 $\pm$ 0.23 \\
& SeM$^2_e$ & 3.59 $\pm$ 0.28 & 3.11 $\pm$ 0.30 & 3.81 $\pm$ 0.27 & 4.21 $\pm$ 0.24 & 3.64 $\pm$ 0.26 \\
\midrule
\multirow{2}{*}{No-Motion} & SeM$^2$ & 3.86 $\pm$ 0.23 & 4.07 $\pm$ 0.20 & 3.54 $\pm$ 0.26 & 4.37 $\pm$ 0.18 & 3.83 $\pm$ 0.22 \\
& SeM$^2_e$ & 3.71 $\pm$ 0.26 & 3.89 $\pm$ 0.24 & 3.38 $\pm$ 0.28 & 4.18 $\pm$ 0.22 & 3.70 $\pm$ 0.25 \\
\midrule
\multirow{2}{*}{Lang-Only} & SeM$^2$ & 2.93 $\pm$ 0.24 & 2.45 $\pm$ 0.26 & N/A & 4.31 $\pm$ 0.20 & 3.02 $\pm$ 0.23 \\
& SeM$^2_e$ & 2.81 $\pm$ 0.27 & 2.32 $\pm$ 0.29 & N/A & 4.12 $\pm$ 0.25 & 2.86 $\pm$ 0.26 \\
\bottomrule
\end{tabular}
}
\end{table*}

\begin{table*}[h!]
\centering
\caption{Human expert evaluation of different system configurations}
\label{tab:human_results}
\resizebox{0.8\textwidth}{!}{%
\renewcommand{\arraystretch}{1.2}
\begin{tabular}{@{}llccccc@{}}
\toprule
\textbf{System} & \textbf{Version} & \textbf{Naturalness} & \textbf{Emotional} & \textbf{Modal} & \textbf{Response} & \textbf{Overall} \\
& & & \textbf{Clarity} & \textbf{Coherence} & \textbf{Appropriateness} & \textbf{Experience} \\
\midrule
\multirow{2}{*}{Full} & SeM$^2$ & 4.52 $\pm$ 0.31 & 4.64 $\pm$ 0.28 & 4.28 $\pm$ 0.32 & 4.36 $\pm$ 0.29 & 4.60 $\pm$ 0.27 \\
& SeM$^2_e$ & 4.35 $\pm$ 0.34 & 4.42 $\pm$ 0.33 & 4.10 $\pm$ 0.36 & 4.18 $\pm$ 0.35 & 4.41 $\pm$ 0.31 \\
\midrule
\multirow{2}{*}{No-Expr} & SeM$^2$ & 3.80 $\pm$ 0.36 & 3.33 $\pm$ 0.38 & 3.88 $\pm$ 0.35 & 4.28 $\pm$ 0.30 & 3.82 $\pm$ 0.34 \\
& SeM$^2_e$ & 3.65 $\pm$ 0.39 & 3.18 $\pm$ 0.42 & 3.72 $\pm$ 0.38 & 4.10 $\pm$ 0.36 & 3.69 $\pm$ 0.37 \\
\midrule
\multirow{2}{*}{No-Motion} & SeM$^2$ & 3.92 $\pm$ 0.35 & 4.15 $\pm$ 0.32 & 3.46 $\pm$ 0.38 & 4.32 $\pm$ 0.28 & 3.90 $\pm$ 0.33 \\
& SeM$^2_e$ & 3.76 $\pm$ 0.38 & 3.94 $\pm$ 0.36 & 3.31 $\pm$ 0.40 & 4.14 $\pm$ 0.34 & 3.75 $\pm$ 0.36 \\
\midrule
\multirow{2}{*}{Lang-Only} & SeM$^2$ & 3.04 $\pm$ 0.37 & 2.56 $\pm$ 0.39 & N/A & 4.25 $\pm$ 0.31 & 3.14 $\pm$ 0.35 \\
& SeM$^2_e$ & 2.88 $\pm$ 0.40 & 2.42 $\pm$ 0.43 & N/A & 4.08 $\pm$ 0.38 & 2.95 $\pm$ 0.39 \\
\bottomrule
\end{tabular}
}
\end{table*}

\section{User Study and Experiment Results}

\subsection{Experiment Materials}


Our experiments were carried out on a humanoid robot platform with 36 degrees of freedom (primarily utilizing the upper body) and a cable-driven flexible hand with 6 degrees of freedom without tactile sensors. For facial expression, we implemented a WS2812 version LED (8$\times$8 pixels) strip controlled by the FASTLED library \footnote{\url{https://github.com/FastLED/FastLED}}, capable of displaying parameterized emotional expressions. The perception system includes an RGB-D camera for facial expression recognition and a microphone array for speech processing. 
For the multimodal generation system, we employed GPT-4o API (version 2024-11) as the teacher model and deployed Minicpm-v2.6 8B \footnote{\url{https://huggingface.co/openbmb/MiniCPM-V}} as our on-device model, which was fine-tuned with a learning rate of 5e-5 for 6 epochs. The software infrastructure is built on ROS2, with optimized inference using PyTorch 2.0.1 and TensorRT 8.6 for real-time performance.

\subsection{Study Design}

Our study was designed to address three primary research questions: \textbf{(1)} Does the multimodal coordination system significantly enhance human-robot interaction compared to unimodal approaches? \textbf{(2)} How effectively does our edge-deployable model maintain the core functionalities of the API-based teacher model? \textbf{(3)} What is the relative contribution of different modalities (facial expression, gestures, and speech) to the overall interaction quality?

AI has been demonstrated to serve as an effective and efficient method for scoring when used as a referee \cite{chen2024humans}.
For evaluation, we implemented a dual assessment framework that combines AI and human evaluation. We recorded 10 interaction videos for each system configuration in each scenario, resulting in 500 total video samples. These videos were assessed using a 5-point Likert scale (1=poor, 5=excellent) measuring five dimensions: interaction naturalness, emotional expression clarity, modal coherence, response appropriateness, and overall user experience. GPT-4V was employed to evaluate all 500 videos, while 37 human experts blindly evaluated 30\% of randomly selected videos to validate AI assessment reliability. The subjects were college students and community residents (not members of the research team). The interaction content covered 5 scenarios, including daily greetings and emotional confession.

Due to the challenges of direct comparison with existing works (e.g., hardware-specific embodiment, limited reproducibility of proprietary systems), we focus on ablation studies and unimodal variants to validate the proposed framework, following similar evaluation paradigms in \cite{mahadevan2024generative}. We will release our code after the paper is accepted.


\begin{figure*}[]
    \centering
    \includegraphics[width=1.0\textwidth]{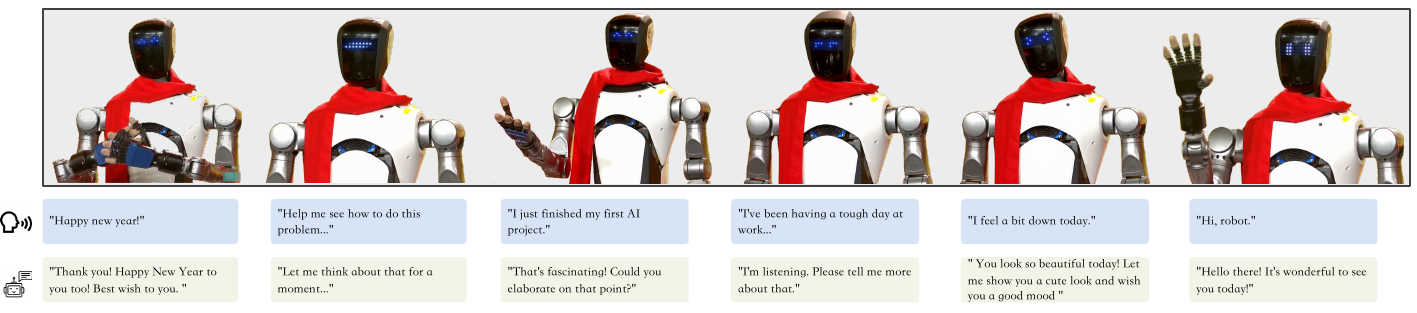}
    \caption{Representative examples of HRI using SeM$^2$ across diverse scenarios. Each column shows a user input and its response.}
    \label{fig:case}
\end{figure*}

\subsection{Main Results}

Our evaluation results, presented in Tables \ref{tab:main_results} and \ref{tab:human_results}, demonstrate the effectiveness of SeM$^2$ across multiple interaction scenarios. Both AI and human evaluations show similar results, validating our evaluation approach.

The complete multimodal system significantly outperformed all reduced variants across all metrics, confirming the superiority of coordinated multimodal expression for human-robot interaction. Particularly noteworthy is the high performance in emotional clarity and overall user experience, where the entire system received the highest scores from AI and human evaluators. The results also highlight a key finding: our edge-deployed 8B model (SeM$^2_e$) maintained approximately 95\% of the server-based teacher model's performance while enabling real-time operation on resource-constrained hardware. This demonstrates that sophisticated multimodal interaction systems can be effectively deployed in practical robotic applications without requiring cloud connectivity or powerful onboard computing resources. Furthermore, the progressive decline in user experience metrics as modalities are removed confirms that rich, multi-channel expression is fundamental to creating natural and expressive human-robot interactions.

\begin{table}[t!]
\centering
\caption{Ablation study of SSAM components}
\label{tab:ablation_results}
\resizebox{0.4\textwidth}{!}{%
\renewcommand{\arraystretch}{1.2}
\begin{tabular}{@{}lccccc@{}}
\toprule
\textbf{Configuration} & \textbf{Nat.} & \textbf{Emot.} & \textbf{Coh.} & \textbf{Resp.} & \textbf{Overall} \\  
\midrule
w. SSAM & 4.42 & 4.51 & 4.38 & 4.45 & 4.53  \\
\midrule
w/o. Modal Sync & 3.65 & 4.23 & 3.27 & 4.36 & 3.74 \textcolor{red}{(-0.79)} \\
w/o. Context Map & 3.98 & 3.86 & 4.15 & 4.31 & 4.07 \textcolor{red}{(-0.46)} \\
w/o. Temporal Plan & 3.76 & 4.32 & 3.52 & 4.40 & 3.89 \textcolor{red}{(-0.64)} \\
w/o. Emotion Intens. & 4.25 & 4.08 & 4.29 & 4.43 & 4.30 \textcolor{red}{(-0.23)} \\
Rule-based & 3.41 & 3.68 & 3.12 & 4.29 & 3.45 \textcolor{red}{(-1.08)} \\
\bottomrule
\multicolumn{6}{l}{\footnotesize Nat.=Naturalness, Emot.=Emotional Clarity, Coh.=Modal Coherence,} \\
\multicolumn{6}{l}{\footnotesize Resp.=Response Appropriateness}
\end{tabular}
}
\end{table}

\begin{table}[t!]
\centering
\caption{Response time (seconds) of two models in various scenarios}
\label{tab:timing_analysis}
\resizebox{0.45\textwidth}{!}{%
\renewcommand{\arraystretch}{1.2}
\begin{tabular}{@{}lcccc@{}}
\toprule
\multirow{2}{*}{\textbf{Scenario}} & \multicolumn{2}{c}{\textbf{First Response}} & \multicolumn{2}{c}{\textbf{Turn Duration}} \\
\cmidrule(lr){2-3} \cmidrule(lr){4-5}
& \textbf{SeM$^2$} & \textbf{SeM$^2_e$} & \textbf{SeM$^2$} & \textbf{SeM$^2_e$} \\
\midrule
Daily Chat (DC) & 5.37 & 2.18 & 10.45 & 9.82 \\
Emotional Talk (ET) & 7.62 & 3.46 & 16.35 & 15.21 \\
Query Answering (QA) & 8.24 & 4.21 & 18.73 & 17.54 \\
Knowledge QA (KQA) & 9.58 & 4.85 & 19.45 & 18.26 \\
\midrule
\textbf{Average} & 7.70 & 3.68 & 16.25 & 15.21 \\
\bottomrule
\end{tabular}
}
\end{table}

\subsection{Ablation Study}

To evaluate the contribution of individual components to SSAM, we conduct an ablation study as shown in Table \ref{tab:ablation_results}. Modal synchronization emerges as the most critical component, with its removal causing the largest decrease in modal coherence and overall experience. Temporal Planning similarly proves essential for naturalness, while Context Mapping primarily influences emotional clarity. The rule-based approach performs substantially worse than any partial SSAM configuration, confirming the advantage of our learning-based architecture over traditional fixed mapping strategies. Interestingly, response appropriateness remains consistent across all configurations, suggesting that semantic content quality is preserved regardless of expression coordination quality, while the expressive dimensions of interaction are significantly enhanced by our complete SSAM framework.

\subsection{Case Study}

To illustrate clearly, we present several example expressions representing different scenarios, as depicted in Figure \ref{fig:case}. The system demonstrates coherent coordination between facial displays and verbal responses, adapting its expression style to fit different contextual needs, from celebratory exchanges to emotional support and problem solving assistance. This semantic-temporal alignment, enabled by our SSAM mechanism, was particularly noted in user feedback, with participants highlighting the natural flow of these multimodal expressions as a key factor in perceived social presence.

\subsection{The Analysis of Response Time}

To evaluate the advantages of edge deployment, response latency is measured in various interaction scenarios, as depicted in Table \ref{tab:timing_analysis}. SeM$^2_e$ achieves approximately 52\% faster first response times compared to the cloud-based solution, while maintaining comparable turn durations in all scenarios tested. Notably, these measurements are acquired specifically in ideal network settings; In real-world scenarios where connectivity varies, the performance difference would likely increase significantly. Besides aiding deployment on resource-limited hardware, the edge model enhances real-time interaction capabilities, vital for maintaining smooth user engagement, since delays in response significantly impact user experience.

\section{CONCLUSION}

In this work, we present SeM$^2$ and edge-deployable SeM$^2_e$, demonstrating that coordinated multimodal expressions significantly enhance HRI naturalness. Through SSAM and knowledge distillation, we establish both the technical feasibility and practical value of sophisticated emotional expression systems on resource-constrained robotic platforms.

Despite our contributions, we acknowledge several limitations, such as concentrating solely on upper-body interactions without integrating locomotion and facing constraints tied to specific platform implementations. Future research directions include expanding to full-body coordination, adapting to diverse robot embodiments, and developing more sophisticated temporal synchronization for extended interactions. These advancements would support the development of truly autonomous socially expressive robots capable of maintaining engaging human interaction across unpredictable real-world environments without relying on cloud connectivity.

\bibliographystyle{IEEEtranN}
\bibliography{root}

\end{document}